\documentclass[11pt]{article}

\usepackage[final]{acl}

\usepackage{times}
\usepackage{latexsym}

\usepackage[T1]{fontenc}

\usepackage[utf8]{inputenc}

\usepackage{microtype}

\usepackage{inconsolata}

\usepackage{graphicx}

\usepackage{graphicx} 
\usepackage{enumitem}
\usepackage{subcaption}
\usepackage{booktabs}
\usepackage{amsfonts}
\usepackage{amsmath}
\usepackage{booktabs}
\usepackage{multirow}
\usepackage{graphicx}
\usepackage{enumitem}
\usepackage{textcomp}
\usepackage{booktabs}

\usepackage{array}
\usepackage{bm}
\usepackage{tabularx}
\usepackage{tabulary}
\usepackage{array}
\usepackage{color,ulem}
\usepackage{xcolor}
\usepackage{caption}
\usepackage{makecell} 
\usepackage{xcolor} 
\usepackage{fontawesome5} 
\usepackage{hyperref}
\usepackage[table]{xcolor}

\usepackage{algorithm}
\usepackage{algorithmic}
\usepackage{array}
\usepackage{bm}
\usepackage{array}
\usepackage{float}  
\usepackage{placeins}

\title{Entropy Ratio Clipping as a Soft Global Constraint \\ for Stable Reinforcement Learning}

\author{
Zhenpeng Su\textsuperscript{\rm 1}\footnotemark[1] \quad Leiyu Pan \textsuperscript{\rm 1}\footnotemark[1] \quad Minxuan Lv\textsuperscript{\rm 1} \quad Tiehua Mei\textsuperscript{\rm 1} \quad Zijia Lin\textsuperscript{\rm 2} \quad Yuntao Li\textsuperscript{\rm 3} \\ \quad  \textbf{Wenping Hu}\textsuperscript{\rm 1} \quad \textbf{Ruiming Tang}\textsuperscript{\rm 1} \quad \textbf{Kun Gai}\textsuperscript{\rm 1} \quad \textbf{Guorui Zhou}\textsuperscript{\rm 1} \footnotemark[2] \\
  \textsuperscript{\rm 1}Kuaishou Technology
  \textsuperscript{\rm 2}Tsinghua University, Beijing, China
  \textsuperscript{\rm 3}Independent\\
  \faEnvelope\ \href{suzhenpeng13@163.com}{suzhenpeng13@163.com}
}

\begin{document}
\maketitle
\renewcommand{\thefootnote}{\fnsymbol{footnote}} 
\footnotetext[1]{Equal contribution. This work was completed by Leiyu Pan during an internship at Kuaishou.}
\footnotetext[2]{Corresponding authors. } 
\renewcommand{\thefootnote}{\arabic{footnote}}
\begin{abstract}


Large language model post-training relies on reinforcement learning to improve model capability and alignment quality. 
However, the off-policy training paradigm introduces distribution shift, which often pushes the policy beyond the trust region, leading to training instabilities manifested as fluctuations in policy entropy and unstable gradients. Although PPO-Clip mitigates this issue through importance clipping, it still overlooks the global distributional shift of actions.
To address these challenges, we propose using the entropy ratio between the current and previous policies as a new global metric that effectively quantifies the relative change in policy exploration throughout updates. Building on this metric, we introduce an \textbf{Entropy Ratio Clipping} (ERC) mechanism that imposes bidirectional constraints on the entropy ratio. This stabilizes policy updates at the global distribution level and compensates for the inability of PPO-clip to regulate probability shifts of un-sampled actions. We integrate ERC into both DAPO and GPPO reinforcement learning algorithms. Experiments across multiple benchmarks show that ERC consistently improves performance. 

\end{abstract}

\section{Introduction}

In the post-training stage of large language models (LLMs), reinforcement learning (RL) has gradually become a core paradigm for improving both capability and alignment quality ~\cite{DBLP:conf/nips/Ouyang0JAWMZASR22,DBLP:journals/corr/abs-2402-03300,DBLP:journals/nature/GuoYZSWZXZMBZY025}. By sampling trajectories and updating policies based on reward signals, models can achieve superior performance on complex reasoning tasks ~\cite{DBLP:journals/corr/abs-2505-09388,DBLP:journals/corr/abs-2504-13914}. Among various RL,  Reinforcement Learning with Verifiable Rewards (RLVR) has recently gained increasing attention, as it enables reward signals to be evaluated in a rule-based manner and significantly enhances the reasoning capability of LLMs ~\cite{DBLP:journals/corr/abs-2411-15124,DBLP:journals/corr/abs-2508-07629}.

However, RL training still faces the persistent challenge of trust-region deviation ~\cite{DBLP:conf/icml/SchulmanLAJM15,DBLP:journals/corr/abs-2505-24864}. Since modern RL for LLMs often adopts an off-policy paradigm, the data used to update the current policy are generated by older behavior policies, leading to distributional drift between the old and new policies. Mainstream methods typically employ importance sampling to correct this bias, yet its inherently high variance can destabilize the update step size ~\cite{DBLP:journals/corr/SchulmanWDRK17}. As a result, policy updates may deviate from the theoretical trust region, triggering a series of training instabilities.

Trust-region deviations readily lead to two problems:

\begin{itemize}[leftmargin=*]
    \item \textbf{Entropy instability}: The policy entropy fluctuates drastically across training stages, leading to excessive or degenerate exploration behavior ~\cite{DBLP:journals/corr/abs-2505-22617,DBLP:journals/corr/abs-2506-14758}.
    \item \textbf{Gradient norm instability}: The gradient magnitude exhibits explosion or vanishing phenomena, impairing convergence and optimization performance ~\cite{DBLP:journals/corr/abs-2505-24864,DBLP:journals/corr/abs-2506-14731}.
\end{itemize}

\begin{figure*}[t]
\centering
\begin{subfigure}[t]{0.32\textwidth}
    \centering
    \includegraphics[height=3.6cm,keepaspectratio]{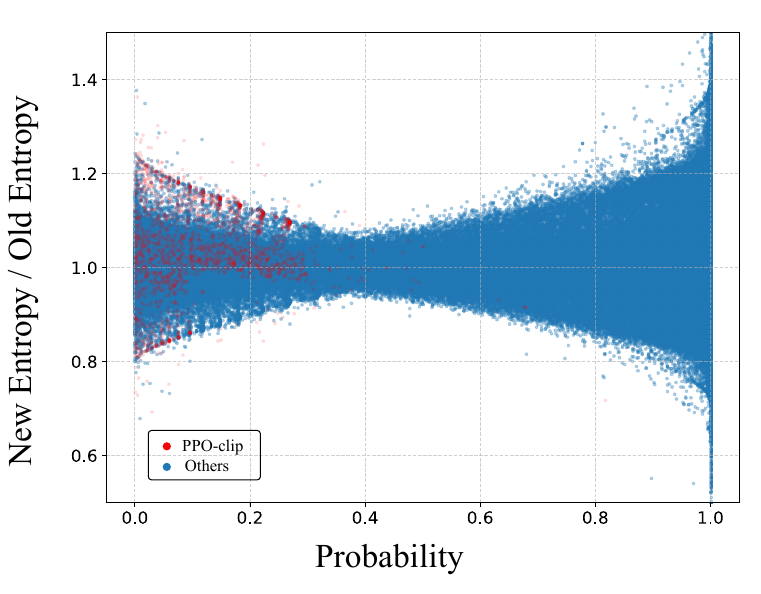}
    \caption{Entropy ratio versus old probability}
    \label{fig:sub1}
\end{subfigure}
\hfill
\begin{subfigure}[t]{0.32\textwidth}
    \includegraphics[height=3.8cm,keepaspectratio]{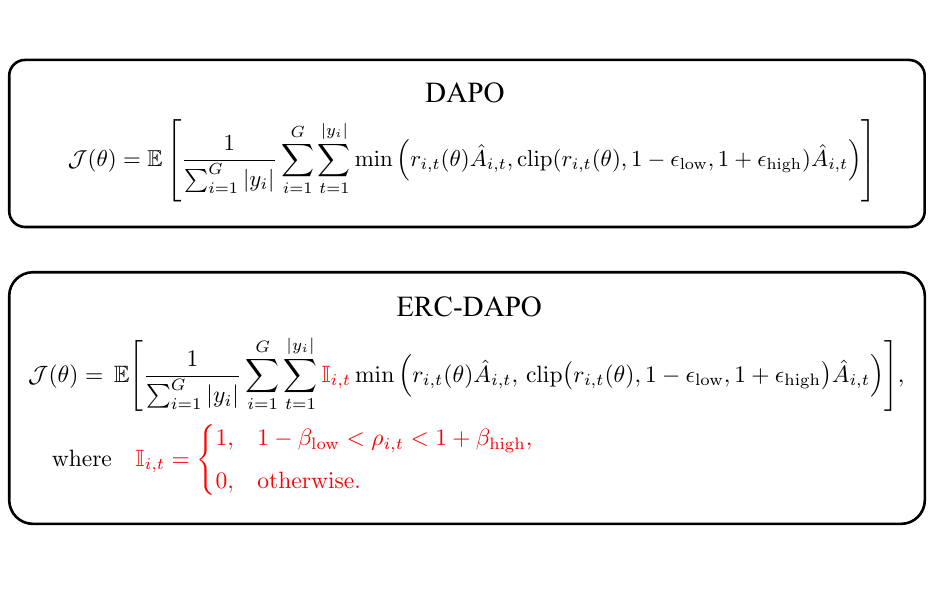}
    \caption{Formulation of ERC}
    \label{fig:sub2}
\end{subfigure}
\hfill
\begin{subfigure}[t]{0.32\textwidth}
    \centering
    \includegraphics[height=3.6cm,keepaspectratio]{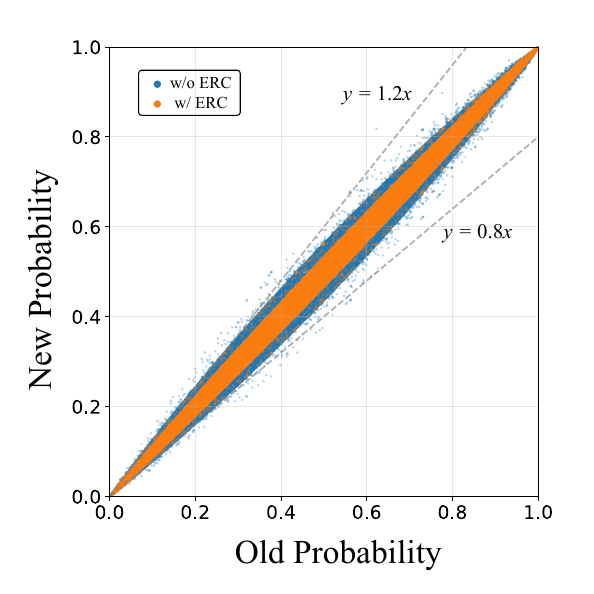}
    \caption{New probability versus old probability}
    \label{fig:sub3}
\end{subfigure}

\caption{\textbf{(a)}: Scatter plot showing the relationship between token-wise sampling probability and entropy ratio during RL training. \textbf{(b)}: Comparison of the optimization objectives for DAPO and DAPO augmented with ERC.
ERC extends the standard PPO-clip objective in DAPO by introducing an additional clipping term on the entropy ratio $\rho_{i,t}$, thereby enforcing a global distribution-level constraint. \textbf{(c)}: Comparison of the trust regions with and without ERC. By applying bidirectional clipping on the entropy ratio, ERC further tightens the trust region beyond PPO-clip, effectively mitigating trust-region drift.}
\label{fig:fig1}
\end{figure*}

Existing works primarily follow two paradigms to ensure the reliability of the trust region~\cite{DBLP:journals/corr/SchulmanWDRK17}. Firstly, PPO-Penalty~\cite{DBLP:journals/corr/SchulmanWDRK17} introduces a KL divergence penalty into the policy gradient objective, using a single coefficient to control the overall divergence between the old and new policies and prevent excessively large updates. However, this coefficient is highly sensitive: an insufficient penalty under-regularizes the optimization, leaving the policy vulnerable to instability; an excessive one over-constrains the parameter space and usually hinders exploration. To address PPO-Clip employs a ``hard clipping'' mechanism that restricts the importance sampling ratio within a predefined interval, preserving exploration capacity while suppressing drastic changes in the sampled actions. Empirical results show that this approach is simple and effective, yet it has a blind spot: the probabilities of unsampled actions remain entirely unconstrained. 

As iterations proceed, this portion of the distribution continues to drift, ultimately threatening policy stability. For example, if the action space is $\mathsf{\{\textit{a}, \textit{b}, \textit{c}, \textit{d}\}}$, the old policy probabilities are $\mathsf{\{0.85, 0, 0.15, 0\}}$, after multiple iterations, the new policy probabilities become $\mathsf{\{0.82, 0.064, 0.07, 0.046\}}$. Although the probability of the sampled action $\mathsf{\textit{a}}$ changes only slightly and PPO-Clip does not trigger clipping, the distribution of the remaining actions has shifted significantly, potentially causing oscillations in subsequent updates. As shown in Figure~\ref{fig:sub1}, when the sampling probability is extremely low or high (e.g., below 0.2 or above 0.6), the global distribution shift becomes more pronounced, particularly for high-probability tokens. In these cases, PPO-clip fails to effectively constrain such significant global deviations, as its clipping primarily occurs on low-probability tokens.

Additionally, previous works have observed that entropy often becomes unstable during PPO-Clip training~\cite{DBLP:journals/corr/abs-2503-14476,DBLP:journals/corr/abs-2509-20712,DBLP:journals/corr/abs-2508-07629}. We argue that one cause lies in the inability to clip actions where entropy changes drastically between the old and new policies. For instance, the entropy of the old policy in the above example is $0.422$, while that of the new policy increases sharply to $0.666$. This unconstrained entropy variation leads to significant fluctuations during training. 

Inspired by PPO-Clip, we propose the Entropy Ratio Clipping (ERC) mechanism. As shown in the Figure~\ref{fig:sub2}, ERC directly applies hard truncation to sample gradients when the entropy change between the old and new policies exceeds an allowable range. ERC does not replace PPO-Clip but complements it: while PPO-Clip only constrains the magnitude of local updates for sampled actions, ERC clamps the entropy ratio within a moderate interval, mitigating the drift of the overall policy distribution. Experiments demonstrate that this hard constraint simultaneously stabilizes both entropy values and gradients throughout the training process, ultimately leading to consistent and significant performance improvements.

Furthermore, as illustrated in Figure~\ref{fig:sub3}, our quantitative analysis demonstrates that incorporating ERC significantly narrows and stabilizes the effective trust region. Even under substantial off-policy conditions, the method with ERC consistently maintains an importance sampling ratio closer to 1 compared to the approach without ERC. Empirically, this results in a more reliable and stable optimization process, reinforcing both convergence consistency and policy robustness.

The main contributions of our work can be summarized as follows:

\begin{itemize}[leftmargin=*]
    \item We introduce the entropy ratio, a novel metric that quantifies the relative change in policy exploration during reinforcement learning training, providing a new dimension for measuring the global drift of policy distributions across updates.
    \item We propose the Entropy Ratio Clipping mechanism, which globally constrains the variation in exploration to effectively mitigate trust-region deviation and enhance training stability.
    \item We integrate and evaluate ERC across multiple reinforcement learning algorithms, demonstrating that our method consistently stabilizes training dynamics and yields performance improvements across a range of benchmarks.
\end{itemize}

\section{Preliminary}

\subsection{Proximal Policy Optimization}

PPO ~\cite{DBLP:journals/corr/SchulmanWDRK17} is one of the most widely adopted policy gradient methods in RL. PPO is to stabilize training by restricting the deviation between the new and old policies during updates, preventing excessively large policy steps. 

Let $\pi_{\text{old}}$ denote the old policy, $\pi_{\theta}$ the current policy, and $A_t$ the advantage function. The standard policy gradient objective can be written as:

\begin{small}
\begin{equation}
\begin{aligned}
\mathcal{J}_{\text{PG}}(\theta) &= 
\mathbb{E}_{x \sim \mathcal{D},\, y \sim \pi_{\theta_{\text{old}}}(\cdot \mid x)} 
\left[ r_t(\theta) A_t \right], \\
r_t(\theta) &= 
\frac{\pi_\theta(y_t \mid x, y_{<t})}
{\pi_{\theta_\text{old}}(y_t \mid x, y_{<t})}
\end{aligned}
\end{equation}
\end{small}

Here, $x$ denotes a query sampled from the data distribution $\mathcal{D}$, 
and $y$ denotes a response generated by the old policy $\pi_{\text{old}}$. 
Directly optimizing this objective may cause the importance ratio $r_t(\theta)$ to deviate excessively, leading to unstable training. To mitigate this issue, PPO introduces two major forms of trust-region constraints.

\paragraph{PPO-penalty}

PPO-penalty enforces a global constraint on the distributional difference between the new and old policies by adding a KL-divergence penalty term to the objective:

\begin{small}
\begin{equation}
\begin{aligned}
\mathcal{J}_{\text{PPO-penalty}}(\theta) =
\mathbb{E} \left[
r_t(\theta) A_t - 
\beta \, \text{KL}\!\left(\pi_{\text{old}} \,\|\, \pi_{\theta}\right)
\right]
\end{aligned}
\end{equation}
\end{small}

Here, $\beta$ is a penalty coefficient. The KL regularizer prevents the new policy from deviating excessively from the old one, thus maintaining training stability. However, PPO-penalty imposes pointwise constraints on every action probability, which may suppress exploration, and the adaptive adjustment of $\beta$ often relies on heuristic or empirical tuning, making stability harder to guarantee.

\paragraph{PPO-clip}

PPO-clip enhances training stability by directly clipping the probability ratio $r_t(\theta)$ within a fixed range, forming a local trust region:

\begin{small}
\begin{equation}
\begin{aligned}
\mathcal{J}_{\text{PPO-clip}}(\theta) =
\mathbb{E} \left[
\min \left(
r_t(\theta) A_t,\,
\text{clip}\!\left(r_t(\theta), 1 - \epsilon, 1 + \epsilon \right) A_t
\right)
\right]
\end{aligned}
\end{equation}
\end{small}

Here, $1-\epsilon$ and $1+\epsilon$ denote the clipping bounds. This mechanism truncates overly large updates to reduce variance and improve stability. Compared with PPO-penalty, PPO-clip is more robust and easier to tune in practice. However, it constrains only sampled actions, leaving unsampled actions unconstrained, which may still drift beyond the trust region.

\subsection{PPO Variants}

\paragraph{Group Relative Policy Optimization (GRPO)}
GRPO ~\cite{DBLP:journals/corr/abs-2402-03300} is a critic-free RL method that simplifies PPO by removing explicit value function estimation. Given a prompt $x$, it estimates advantages by standardizing rewards across a group of $G$ sampled responses $\{r_i\}_{i=1}^G$:

\begin{small}
\begin{equation}
\hat{A}_{i,t} = \frac{r_i - \text{mean}(\{r_i\}_{i=1}^G)}{\text{std}(\{r_i\}_{i=1}^G)}
\end{equation}
\end{small}

The standardized advantages are then applied in a clipped policy gradient objective:

\begin{small}
\begin{equation}
\begin{aligned}
\mathcal{J}_{\text{GRPO}}(\theta) = \mathbb{E} \left[ \frac{1}{G} \sum_{i=1}^{G} \frac{1}{|y_i|} \sum_{t=1}^{|y_i|} \min \big( r_{i,t}(\theta) \hat{A}_{i,t}, \right.\\
\left. \text{clip}(r_{i,t}(\theta), 1-\epsilon, 1+\epsilon) \hat{A}_{i,t} \big) \right]
\end{aligned}
\end{equation}
\end{small}

\paragraph{Decoupled Clip and Dynamic Sampling Policy Optimization (DAPO)}
Building on GRPO, DAPO ~\cite{DBLP:journals/corr/abs-2503-14476} enhances training stability and exploration efficiency through some key modifications. Its optimization objective is as follow:

\begin{small}
\begin{equation}
\begin{aligned}
\mathcal{J}_{\text{DAPO}}(\theta) = \mathbb{E} \left[ \frac{1}{\sum_{i=1}^G |y_i|} \sum_{i=1}^G \sum_{t=1}^{|y_i|} \min \left( r_{i,t}(\theta) \hat{A}_{i,t}, \right. \right. \\
\left. \left. \text{clip}(r_{i,t}(\theta), 1-\epsilon_{\text{low}}, 1+\epsilon_{\text{high}}) \hat{A}_{i,t} \right) \right]
\end{aligned}
\end{equation}
\end{small}

Compared with GRPO, DAPO introduces three improvements: asymmetric clipping bounds $(1-\epsilon_{\text{low}}, 1+\epsilon_{\text{high}})$ to encourage exploration; dynamic sample filtering to discard uninformative responses; token-level loss aggregation with reward shaping to better handle variable-length outputs.

\paragraph{Gradient-Preserving Clipping Policy Optimization (GPPO)}
While GRPO and DAPO improve efficiency and stability, the traditional clipping mechanism can still suppress gradients of high-entropy tokens and slow the convergence of negative samples. To address this, \citet{DBLP:journals/corr/abs-2508-07629} proposed GPPO, which preserves gradients when the importance sampling ratio exceeds the clipping range. By maintaining constant-scale updates, GPPO stabilizes training while alleviating excessive gradient truncation. The objective is as follow:

\begin{small}
\begin{equation}
\begin{aligned}
\mathcal{J}_{\text{GPPO}}(\theta) = \mathbb{E} \left[ \frac{1}{\sum_{i=1}^G |y_i|} \sum_{i=1}^G \sum_{t=1}^{|y_i|} \min \left( r_{i,t}(\theta) \hat{A}_{i,t}, \right. \right. \\
\left. \left. \text{clip}(r_{i,t}(\theta), \frac{1-\epsilon_{\text{low}}}{\text{sg}(r_{i,t}(\theta))}r_{i,t}(\theta), \frac{1+\epsilon_{\text{high}}}{\text{sg}(r_{i,t}(\theta))}r_{i,t}(\theta)) \hat{A}_{i,t} \right) \right]
\end{aligned}
\end{equation}
\end{small}

\section{Method}

\subsection{Entropy Ratio}






In RL, off-policy updates often deviate from the trust region, leading to instability during training. Although PPO-clip mitigates excessively large updates by clipping the importance sampling ratio, its constraint applies only to the sampled actions and thus fails to capture the overall change in the policy distribution. To further enhance training stability, we aim to introduce a more comprehensive distributional constraint on top of PPO-clip, while preserving sufficient exploration capability for stable learning. 

To this end, we propose the entropy ratio, defined as the relative change in entropy between the new and old policies evaluated on the same data. Specifically, for each decoding step $t$, the token-level entropy ratio is defined as:

\begin{small}
\begin{equation}
\begin{aligned}
\rho_t &= \frac{\mathcal{H}(\pi_\theta, t)}{\mathcal{H}(\pi_{\text{old}}, t)} \\
&= \frac{-\sum_{a \in \mathcal{V}} \pi_\theta(a \mid y_{<t}, x)\,\log \pi_\theta(a \mid y_{<t}, x)}
{-\sum_{a \in \mathcal{V}} \pi_{\text{old}}(a \mid y_{<t}, x)\,\log \pi_{\text{old}}(a \mid y_{<t}, x)}.
\end{aligned}
\end{equation}
\end{small}
where $\mathcal{V}$ denotes the vocabulary and $a$ represents every token in $\mathcal{V}$. 
Crucially, the entropy ratio overcomes a key limitation of importance sampling, which focuses only on sampled actions, by directly measuring shifts across the entire action distribution, including unsampled actions.





\subsection{Entropy Ratio Clip}


After introducing the entropy ratio as a global constraint on the policy distribution, we further incorporate this constraint into existing reinforcement learning objectives.
Inspired by PPO-clip, we propose the Entropy Ratio Clipping (ERC) mechanism, which discards gradients of tokens whose entropy ratio $\rho_t$ falls outside the predefined range $(1-\beta_{\text{low}}, 1+\beta_{\text{high}})$.
Taking DAPO as an example, the ERC objective can be formalized as follows:

\begin{small}
\begin{equation}
\begin{aligned}
& \mathcal{J}_{\text{ERC}}(\theta)
= \, \mathbb{E} \Bigg[
\frac{1}{\sum_{i=1}^{G}\left|y_{i}\right|}
\sum_{i=1}^{G} \sum_{t=1}^{\left|y_{i}\right|}
\mathbb{I}_{i,t} \\
& \min \Big(
r_{i,t}(\theta) \hat{A}_{i,t},\,
\text{clip}\big(
r_{i,t}(\theta),
1-\epsilon_{\text{low}},
1+\epsilon_{\text{high}}
\big)\hat{A}_{i,t}
\Big)
\Bigg], \\
& \quad \text{where} \quad 
\mathbb{I}_{i,t} =
\begin{cases}
1, & 1- \beta_{\text{low}} < \rho_{i,t} < 1 + \beta_{\text{high}}, \\[4pt]
0, & \text{otherwise.}
\end{cases}
\end{aligned}
\end{equation}
\end{small}


If an update causes the entropy ratio to exceed its preset range, ERC directly applies a hard truncation to the corresponding gradients, preventing sharp fluctuations in the global output distribution and entropy. Unlike KL constraints that continuously restrict the policy throughout training, the entropy ratio becomes active only when the entropy of the new policy is about to deviate substantially from that of the old policy. This approach prevents sudden collapses of the policy distribution while preserving sufficient exploration capacity. 

Building upon PPO-Clip, further introducing the ERC to measure the distribution shift between the old and new policies offers two key benefits. First, it can address the issue of global distribution shift caused by importance sampling, which only considers the probability of the sampled actions while ignoring the distribution changes of the unsampled actions. Second, by clipping samples where the entropy ratio deviates significantly, we can more easily maintain stable entropy between the old and new policies.

Experiments show that compared with PPO-clip, this constraint stabilizes the entropy curve, reduces gradient variance, and enables the model to perform conservative updates while maintaining ongoing exploration, ultimately achieving more stable and efficient policy optimization. In practice, the ERC mechanism integrates orthogonally with various reinforcement learning objectives that rely on importance-ratio clipping.

\section{Experiment}

\begin{figure}[t]
    \centering
    \begin{subfigure}{0.48\linewidth}
        \centering
        \includegraphics[width=\linewidth]{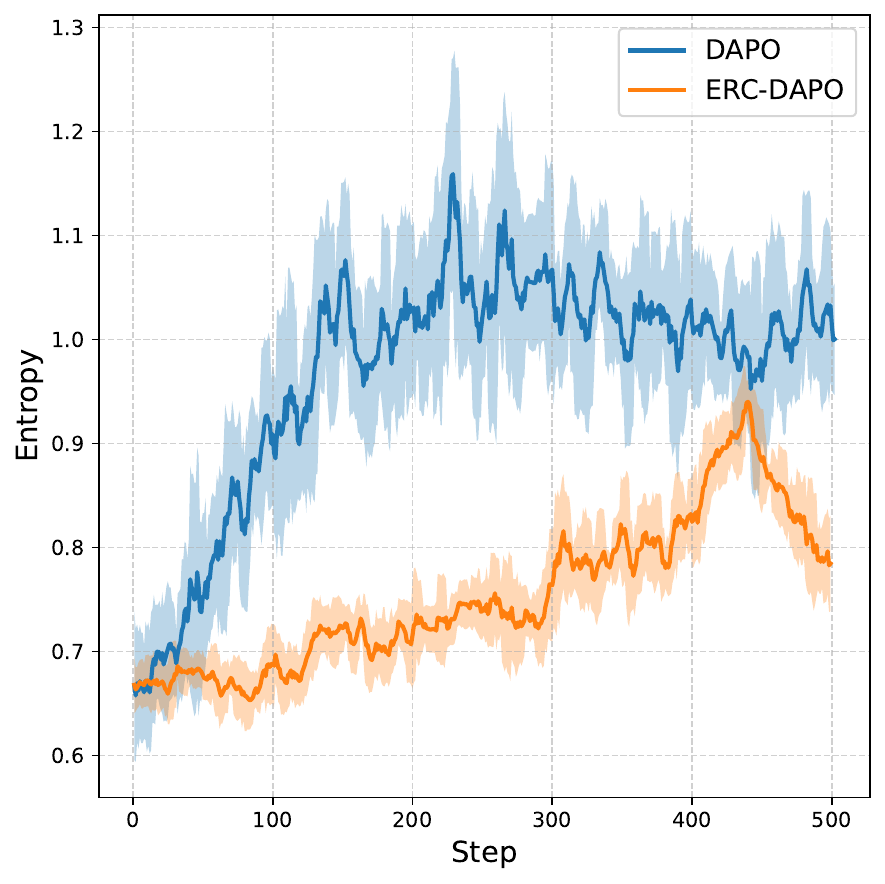}
        \caption{Entropy}
        \label{fig:entropy}
    \end{subfigure}
    \hfill
    \begin{subfigure}{0.48\linewidth}
        \centering
        \includegraphics[width=\linewidth]{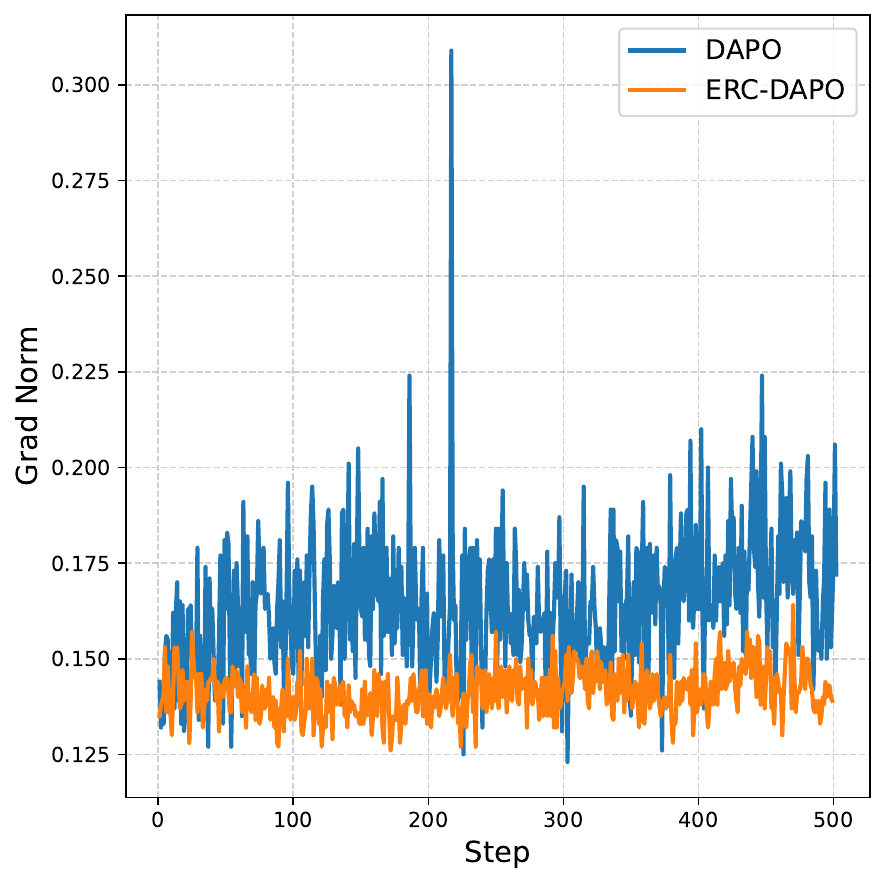}
        \caption{Gradient Norm}
        \label{fig:grad_norm}
    \end{subfigure}
    
    \begin{subfigure}{0.48\linewidth}
        \centering
        \includegraphics[width=\linewidth]{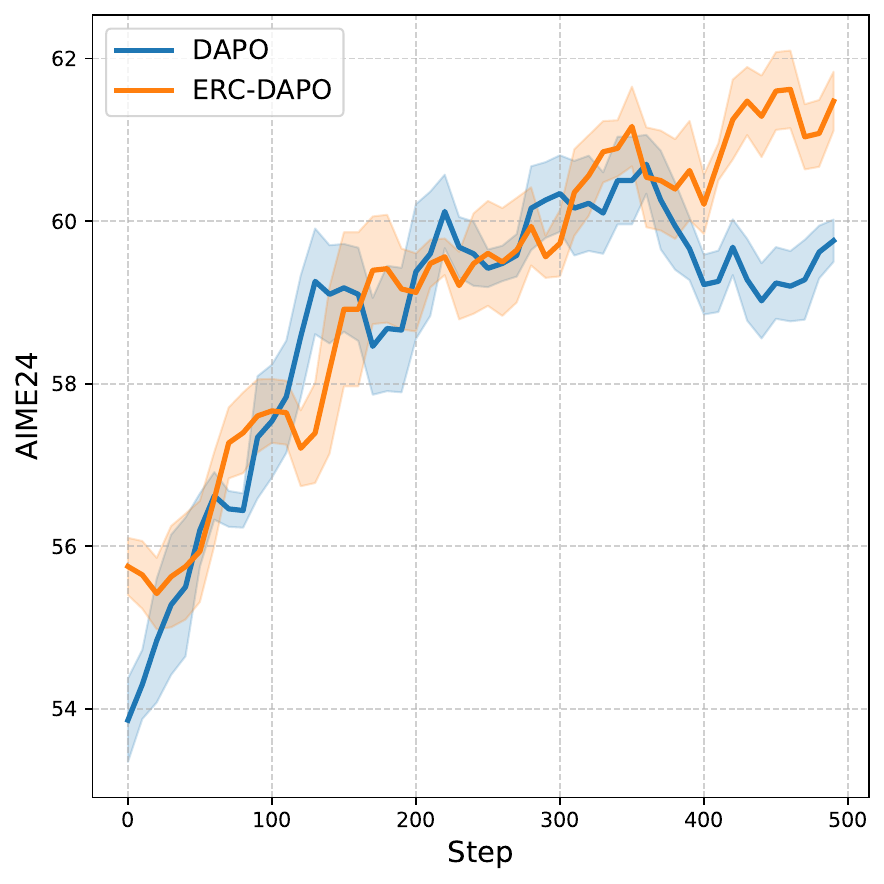}
        \caption{AIME24}
        \label{fig:aime24}
    \end{subfigure}
    \hfill
    \begin{subfigure}{0.48\linewidth}
        \centering
        \includegraphics[width=\linewidth]{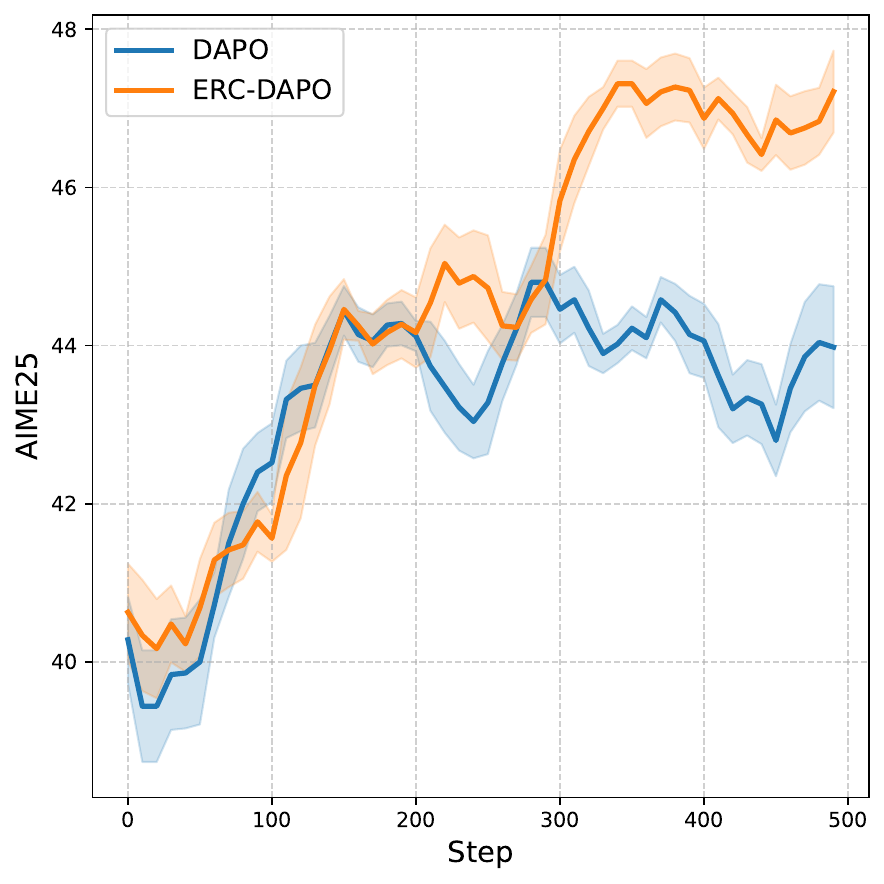}
        \caption{AIME25}
        \label{fig:aime25}
    \end{subfigure}
    
    \caption{Training dynamics of entropy, gradient norm and benchmark accuracy on DeepSeek-R1-Distill-Qwen-7B, comparing various baseline method with and without the proposed ERC mechanism.}
    \label{fig:fig2}
\end{figure}

\subsection{Experimental Setup}

\paragraph{Datasets}


Our training data is derived from the KlearReasoner-MathSub-30K dataset ~\cite{DBLP:journals/corr/abs-2508-07629}, which contains 30k high-quality mathematical reasoning samples. This dataset integrates multiple curated sources, including Skywork-OR1 ~\cite{DBLP:journals/corr/abs-2505-22312}, Acereason ~\cite{DBLP:journals/corr/abs-2505-16400}, NuminaMath ~\cite{numina_math_datasets}, and DeepScaleR ~\cite{deepscaler2025}, followed by rigorous filtering and data decontamination. Specifically, for each query, we distilled 16 responses using DeepSeek-R1-0120 and retained only those queries for which the majority of responses passed rule-based validators \texttt{math-verify}\footnote{https://github.com/huggingface/Math-Verify}. This ensures both the correctness and difficulty of the dataset.

\paragraph{Training}


We trained our models based on two scales of pretrained models: DeepSeek-R1-Distill-Qwen-1.5B\footnote{https://huggingface.co/deepseek-ai/DeepSeek-R1-Distill-Qwen-1.5B} and DeepSeek-R1-Distill-Qwen-7B\footnote{https://huggingface.co/deepseek-ai/DeepSeek-R1-Distill-Qwen-7B}. The maximum response sequence length was set to 16k tokens, and the learning rate was 1e-6. For each query, we rolled out 8 sampled responses. Training proceeds off-policy with a batch of 128 prompts; at every model update this batch is split into mini-batches of size 16.
For the DAPO baseline, we set the clipping thresholds to $\epsilon_{\text{low}}=0.2$ and $\epsilon_{\text{high}}=0.28$ ~\cite{DBLP:journals/corr/abs-2503-14476}; for the GPPO baseline, both thresholds were set to 0.2 ~\cite{DBLP:journals/corr/abs-2509-20712}.
Based on the observations from Figure \ref{fig:fig1}, we intentionally adopted an aggressive clipping strategy by selecting the narrowest region of the entropy-ratio distribution as the preservation interval. As a result, the entropy ratio bounds were set to $\beta_{\text{low}}=0.05$ and $\beta_{\text{high}}=0.05$.
\paragraph{Evaluation}



We assessed our approach across a range of open-ended benchmarks, including mathematical reasoning, code reasoning and instruction-following tasks. On the mathematics benchmarks, results were reported using avg@32, except for MATH500, which employed avg@4. For AIME 24/25, inference was performed with a maximum sequence length of 32k, whereas all other mathematics benchmarks were evaluated with a 16k maximum sequence length. Consistent with \citet{yang2024qwen2}, answers were extracted from the model outputs using the \verb|\boxed{}| notation.
For both code reasoning and instruction-following benchmarks, we used avg@4 scores and a 32k maximum inference length. Specifically, for instruction-following tasks, results were reported at both the prompt level (IFBench-P) and the instance level (IFBench-I) to provide a comprehensive evaluation.

\subsection{Main Results}

\begin{table*}[t]
\centering
\resizebox{1.0\linewidth}{!}{ 
\begin{tabular}{lcccccccccc}
\toprule
\multirow{2}{*}{\textbf{Method}} 
& \multicolumn{6}{c}{\textbf{Math Reasoning}} 
& \multicolumn{2}{c}{\textbf{Code Reasoning}} 
& \multicolumn{2}{c}{\textbf{Instruct Following}} \\
\cmidrule(lr){2-7}\cmidrule(lr){8-9}\cmidrule(lr){10-11}
& \textbf{AIME24} & \textbf{AIME25} & \textbf{HMMT25} & \textbf{MATH500} & \textbf{AMC23} & \textbf{Olympiad} & \textbf{HumanEval} & \textbf{LCB v6} & \textbf{IFBench-P} & \textbf{IFBench-I} \\
\midrule
\textbf{DS-R1-Distill-Qwen-1.5B} & 29.2 & 24.1 & 13.1 & 86.0 & 73.7 & 51.8 & 70.4 & 25.1 & 12.0 & 14.1 \\
\quad + GRPO  & 33.4 & 28.1 & 16.6 & 88.3 & 79.3 & 56.2 & 71.5 & 26.6 & 13.4 & 15.8 \\
\quad + DAPO  & 42.0 & 30.3 & 17.6 & 89.4 & 82.3 & 58.6 & 72.3 & 28.5 & 13.2 & 15.1 \\
\rowcolor{gray!15}
\quad + ERC-DAPO  & \textbf{44.2} & \textbf{31.8} & \textbf{19.2} & \textbf{90.0} & \textbf{84.3} & \textbf{61.0} & \textbf{74.0} & \textbf{28.8} & \textbf{13.5} & \textbf{15.9} \\
\midrule
\textbf{DS-R1-Distill-Qwen-7B} & 54.5 & 39.1 & 26.2 & 93.6 & 90.6 & 67.0 & 89.6 & 49.0 & 16.8 & 18.9 \\
\quad + GRPO  & 55.3 & 40.3 & 24.5 & 93.7 & 88.8 & 65.6 & 90.2 & 48.5 & 15.0 & 17.2 \\
\quad + DAPO  & 62.0 & 45.9 & 27.4 & 94.1 & \textbf{92.3} & 69.9 & 91.2 & 50.6 & 15.6 & 18.0 \\
\rowcolor{gray!15}
\quad + ERC-DAPO  & \textbf{62.1} & \textbf{48.4} & \textbf{28.7} & \textbf{95.1} & 91.9 & \textbf{70.9} & \textbf{91.3} & \textbf{51.2} & \textbf{17.7} & \textbf{20.3} \\
\bottomrule
\end{tabular}
}
\caption{Performance comparison of different baselines and ERC-augumented DAPO method on various mathematical reasoning benchmarks. DS-R1-Distill-Qwen-1.5B and DS-R1-Distill-Qwen-7B denote the DeepSeek-R1-Distill-Qwen-1.5B and
DeepSeek-R1-Distill-Qwen-7B models, respectively.}
\label{tab:table1}
\end{table*}

\paragraph{Benchmark Performance}


As shown in Table \ref{tab:table1}, we conducted a comprehensive evaluation of the proposed ERC method across multiple mathematical reasoning benchmarks. Experimental results demonstrate that, compared to existing RL baselines, integrating ERC consistently improves model performance across nearly all benchmarks. Notably, the gains are more pronounced on more challenging benchmarks such as AIME25 and HMMT25, highlighting the strong potential of ERC in complex reasoning scenarios. Moreover, the method yields consistent improvements on both 1.5B and 7B parameter scales, further confirming its robustness and scalability across different model capacities.

\paragraph{Training Stability}




To further investigate the impact of ERC on training dynamics, we compare the evolution of entropy and gradient norms under different methods. As shown in Figure ~\ref{fig:fig2}, traditional clipping methods often exhibit large entropy fluctuations and unstable gradients during training. This instability arises because their constraints apply only to the locally sampled actions, failing to effectively regulate the drift of unsampled actions within the policy distribution. As training progresses, this unconstrained distributional drift leads to trust-region violations and undermines training stability. In contrast, ERC introduces a global entropy-ratio constraint that effectively suppresses global drift in the policy distribution and structurally prevents large entropy shifts during policy updates. As a result, the training process becomes smoother, with more stable entropy trajectories and well-bounded gradient norms.




\section{Analysis}

\subsection{ERC Enhances Trust Region Constraints}


\begin{figure}[t]
    \centering
    \includegraphics[width=0.3\textwidth]{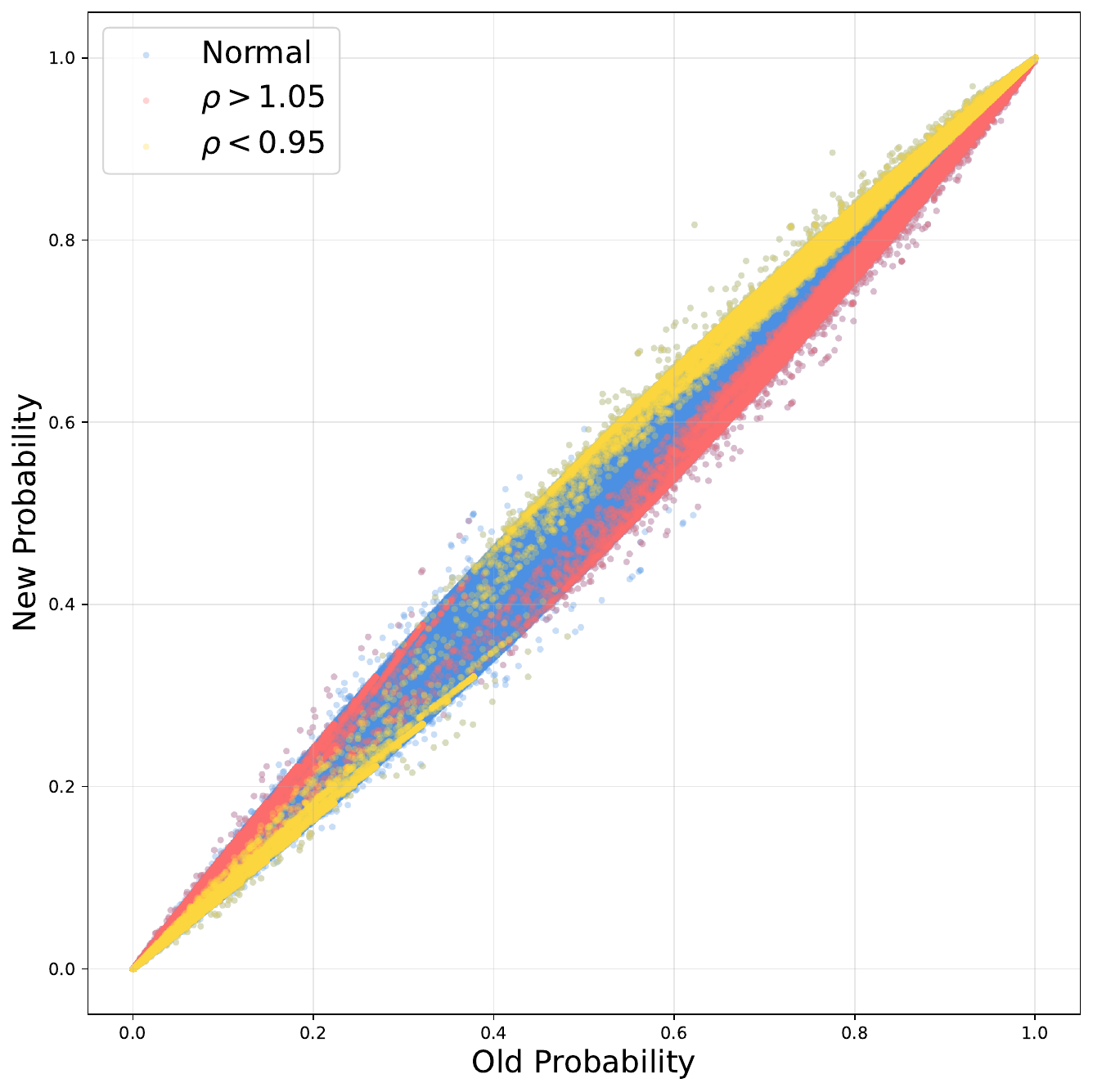}
    \caption{Visualization of the clipping regions. Red points indicate tokens clipped for exceeding the upper bound of the entropy ratio, while yellow points indicate tokens clipped for falling below the lower bound. Blue points represent tokens that were not clipped. The entropy ratio clipping shown here is applied on top of the standard importance ratio clipping.}
    \label{fig:trust_region}
\end{figure}

As shown in Figure \ref{fig:trust_region}, the clipping mechanism of ERC effectively strengthens the trust region constraint. Specifically, the tokens clipped by the entropy ratio boundary are predominantly located near the boundaries of the trust region. This indicates that ERC, operating from a global distribution perspective, can identify and restrict updates to tokens that may still cause the policy to deviate, although overlooked by PPO-clip’s local constraints. Consequently, ERC and PPO-Clip function in a complementary manner, jointly mitigating trust-region divergence and enhancing training stability.

A further analysis of the distribution of clipped tokens reveals that they are mainly concentrated in both high- and low-probability regions. Moreover, the distributions of tokens clipped by the upper and lower bounds exhibit an approximately centrosymmetric pattern. This occurs because a sharp decrease in the probability of high-likelihood tokens or a sharp increase in that of low-likelihood tokens leads to a sudden rise in entropy, triggering clipping by the upper bound. Conversely, the opposite trend causes a sharp entropy decrease and results in clipping by the lower bound. Through this mechanism, ERC effectively restrains drastic fluctuations in the policy distribution.

\subsection{Maintaining Exploration through ERC}
\label{section_5.2}

\begin{figure}[t]
    \centering
    \includegraphics[width=0.4\textwidth]{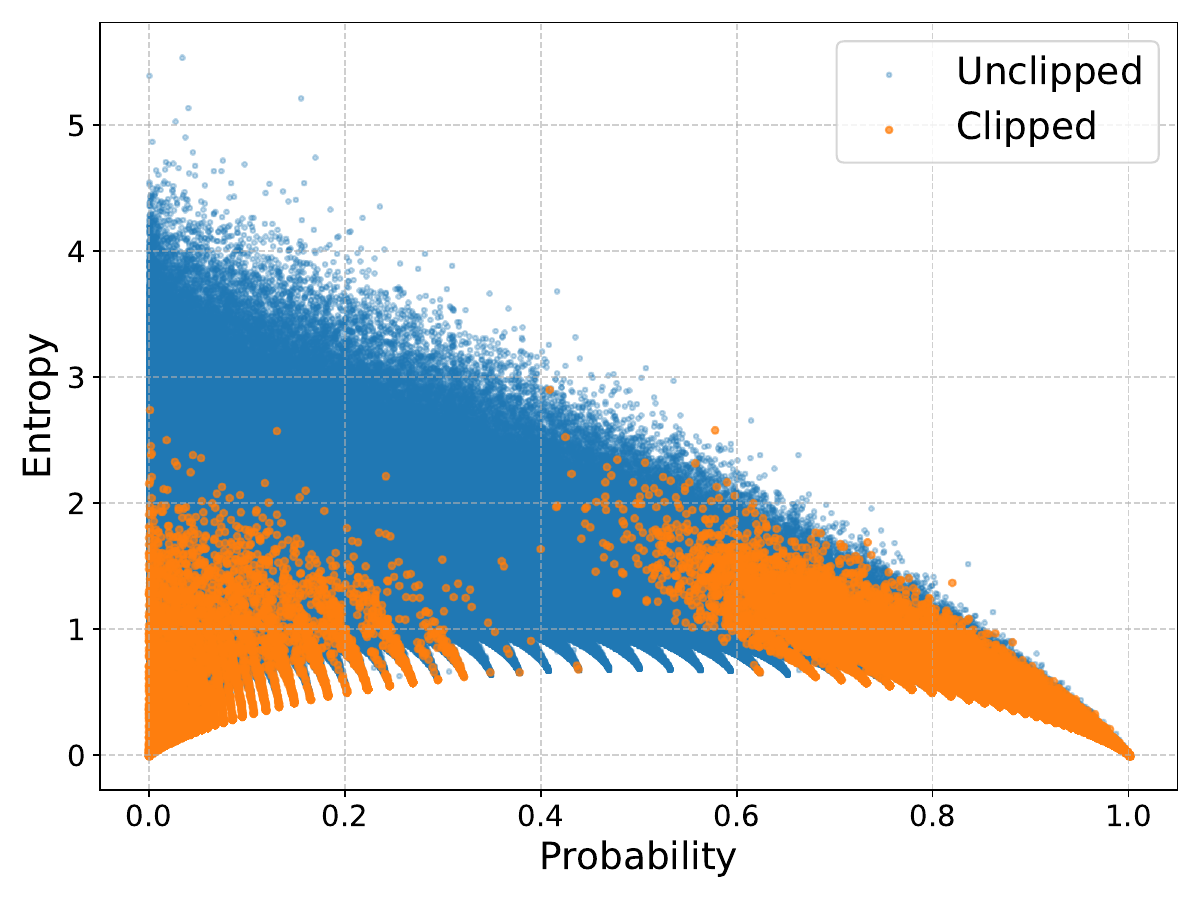}
    \caption{Scatter plot illustrating the relationship between sampled token probabilities and the entropy of their corresponding distributions. Blue points represent tokens that are not clipped by the ERC mechanism, while orange points denote tokens that are clipped by the entropy ratio constraint.}
    \label{fig:entropy}
\end{figure}

\begin{figure}[t]
    \centering
    \begin{subfigure}{0.48\linewidth}
        \centering
        \includegraphics[width=\linewidth]{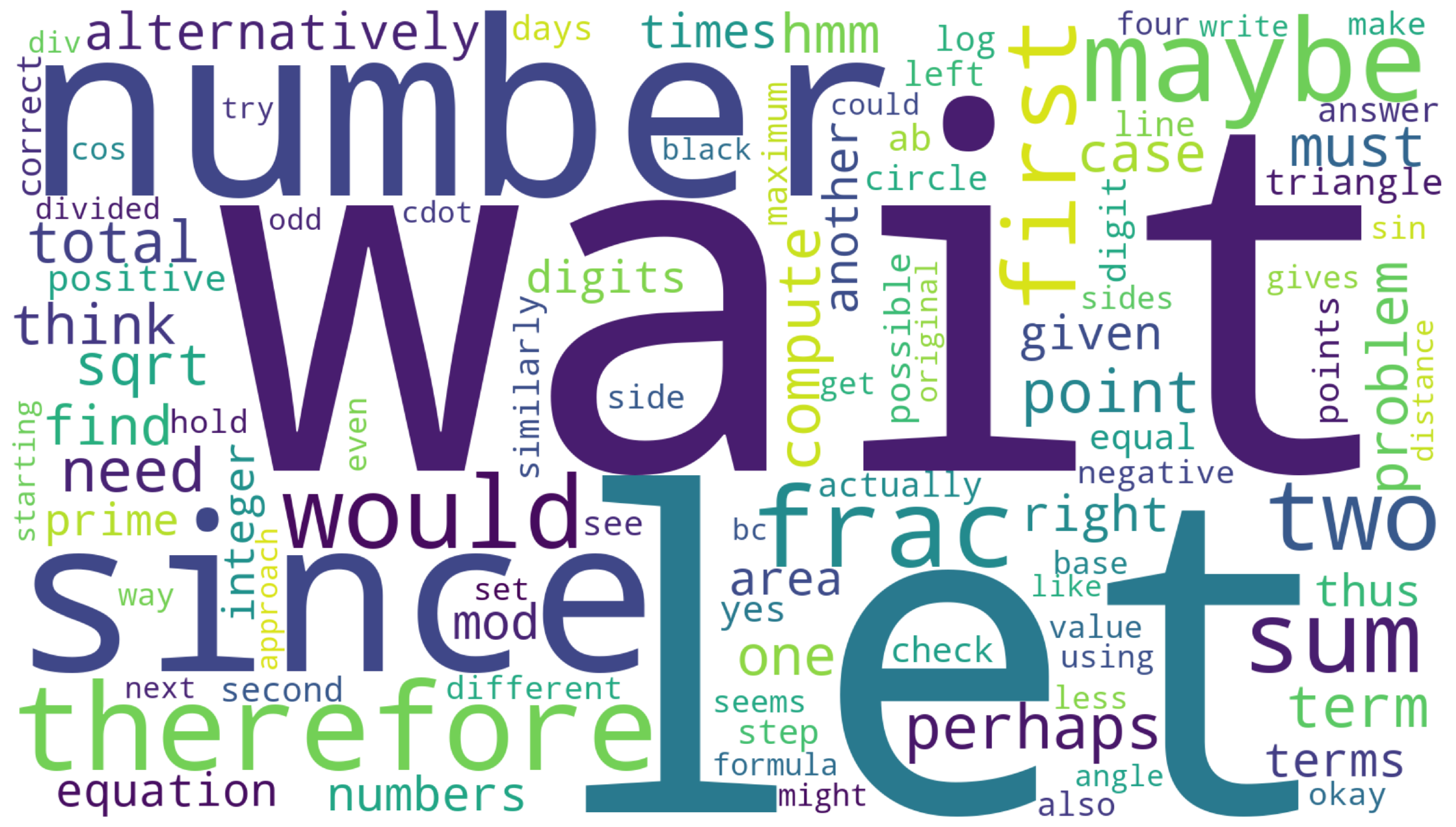}
        \caption{Unclipped Tokens}
        \label{fig:unclipp_token}
    \end{subfigure}
    \hfill
    \begin{subfigure}{0.48\linewidth}
        \centering
        \includegraphics[width=\linewidth]{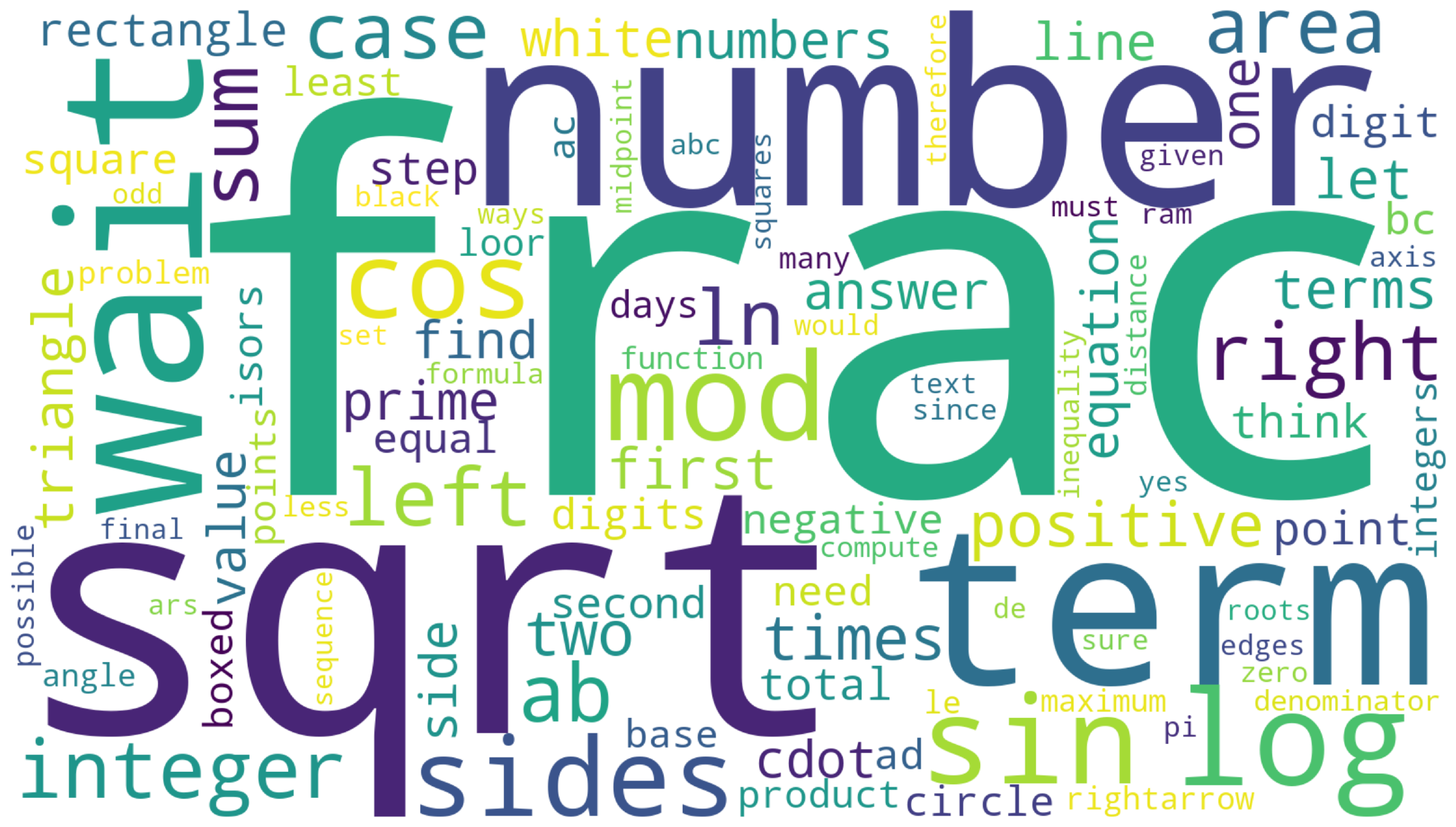}
        \caption{Clipped Tokens}
        \label{fig:clip_token}
    \end{subfigure}
    \caption{Word cloud visualization of tokens unclipped by and clipped by the ERC mechanism.}
\label{fig:word_cloud}
\end{figure}


To further understand the impact of ERC on the model’s exploratory behavior, we analyze the entropy distribution of tokens that are clipped by the entropy ratio constraint during training. As shown in Figure \ref{fig:entropy}, most tokens clipped by ERC fall within low-entropy regions, while high-entropy tokens are generally preserved throughout optimization. This indicates that ERC preferentially suppresses updates to tokens that are overly deterministic and contribute limited information gain, without excessively constraining the model’s exploratory dynamics.

To illustrate this phenomenon more intuitively, we visualize which tokens are clipped and which are retained. As shown in Figure \ref{fig:word_cloud}, the retained tokens often include reasoning-related words such as ``wait'' and ``therefore'' which typically appear in the model’s chain-of-thought and reflect its reasoning exploration process. In contrast, the tokens clipped by ERC are primarily deterministic mathematical symbols or computation operators, such as ``frac'' or ``sqrt'', which contribute little to the diversity of the overall policy distribution.

In summary, ERC not only enforces trust-region constraints but also selectively preserves exploratory updates. This clipping mechanism allows the model to maintain stability while continuing to explore high-entropy decision spaces, achieving a balanced trade-off between training stability and exploratory capability in reinforcement learning.



\subsection{Clipping Ratio Analysis}

\begin{table}[h]
\centering
\begin{tabular}{lc}
\toprule
\textbf{Algorithm} & \textbf{Clip Ratio} \\
\midrule
PPO-clip          & 0.02\% \\
ERC               & 20.29\% \\
\bottomrule
\end{tabular}
\caption{Comparison of token clipping ratios between PPO-clip and ERC.}
\label{tab:clip_ratio_comparison}
\end{table}




Our experimental results show that the global distribution constraint introduced by ERC substantially increases the effective clipping rate. As shown in Table \ref{tab:clip_ratio_comparison}, the clipping ratio under PPO-clip typically remains around 0.02\%, whereas ERC raises this number by nearly three orders of magnitude, reaching approximately 20\%. This striking discrepancy stems from the fundamental difference between the two constraint mechanisms: PPO-clip only regulates the importance ratios of locally sampled actions, where out-of-bound cases are inherently rare; in contrast, ERC extends beyond this local constraint to incorporate a global distributional signal via entropy ratios, enabling it to identify and prune a much larger set of token updates that deviate from the trust region at the distribution level.

Despite ERC’s substantially higher clipping ratio, it consistently surpasses the PPO-clip baselines in both final performance and training stability. This seemingly counterintuitive outcome reveals a key insight: ERC predominantly removes noisy updates that would destabilize training. As discussed in Section \ref{section_5.2}, most tokens clipped by ERC cluster in low-entropy regions, indicating that ERC suppresses overly deterministic and potentially harmful updates while preserving the model’s exploratory behavior elsewhere. This suggests that the truly beneficial training signal in RL is often sparse,  a principle also reflected in methods such as GSPO~\cite{DBLP:journals/corr/abs-2507-18071}, where extensive clipping leads to improved results. Both phenomena reinforce the importance of selectively filtering token-level updates during policy optimization.


\begin{table*}[t]
\centering
\resizebox{0.9\linewidth}{!}{ 
\begin{tabular}{lccccccc}
\toprule
\textbf{Method} & \textbf{AIME24} & \textbf{AIME25} & \textbf{HMMT25} & \textbf{MATH500} & \textbf{AMC23} & \textbf{Olympiad} & \textbf{Avg.} \\
\midrule
\textbf{DS-R1-Distill-Qwen-7B} & 54.5 & 39.1 & 26.2 & 93.6 & 90.6 & 67.0 & 61.8 \\
\quad + GPPO & 57.3 & 46.5 & 24.0 & \textbf{94.7} & 92.0 & 69.9 & 64.1 \\
\rowcolor{gray!15} 
\quad + ERC-GPPO & \textbf{63.5} & \textbf{47.6} & \textbf{28.0} & 94.6 & \textbf{93.5} & \textbf{70.9} & \textbf{66.3} \\
\bottomrule
\end{tabular}
}
\caption{Performance comparison of GPPO and its ERC variant on various mathematical reasoning benchmarks.}
\label{tab:gppo_performance}
\end{table*}

\subsection{The Broader Applicability of ERC}



In our main experimental results, we compared DAPO~\cite{DBLP:journals/corr/abs-2503-14476} with its ERC-augmented variant (ERC-DAPO), demonstrating the effectiveness of integrating ERC into the standard DAPO framework. To further validate the broader applicability of ERC, we additionally combined it with the GPPO method~\cite{DBLP:journals/corr/abs-2508-07629}. 

It is important to highlight the conceptual differences between these two settings. DAPO employs the standard PPO-clip mechanism, in which the gradients of tokens whose importance ratios exceed the clipping bounds are completely discarded. Under this regime, ERC primarily acts as a complementary constraint, compensating for the fact that PPO-clip only regulates locally sampled actions and therefore provides limited coverage over the global policy distribution.
In contrast, GPPO does not rely on standard PPO-clip mechanism. Even when the importance ratio lies outside the clipping interval, GPPO still retains non-zero gradients for those tokens. In this scenario, ERC plays a more central role by serving as the primary stability constraint. Notably, ERC improves performance in both regimes, whether paired with PPO-clip (DAPO) or with a non-clipping method (GPPO).

As shown in Table \ref{tab:gppo_performance}, incorporating ERC into GPPO also yields consistent performance improvements, providing strong evidence for the general effectiveness of ERC across diverse RL algorithms.
These result indicate that ERC is not merely a supplementary component to existing importance-ratio clipping techniques, but also holds the potential to function as an independent and robust constraint mechanism for stabilizing policy optimization.

\subsection{ERC vs. KL Regularization}

\begin{figure}[t]
    \centering
    \begin{subfigure}{0.48\linewidth}
        \centering
        \includegraphics[width=\linewidth]{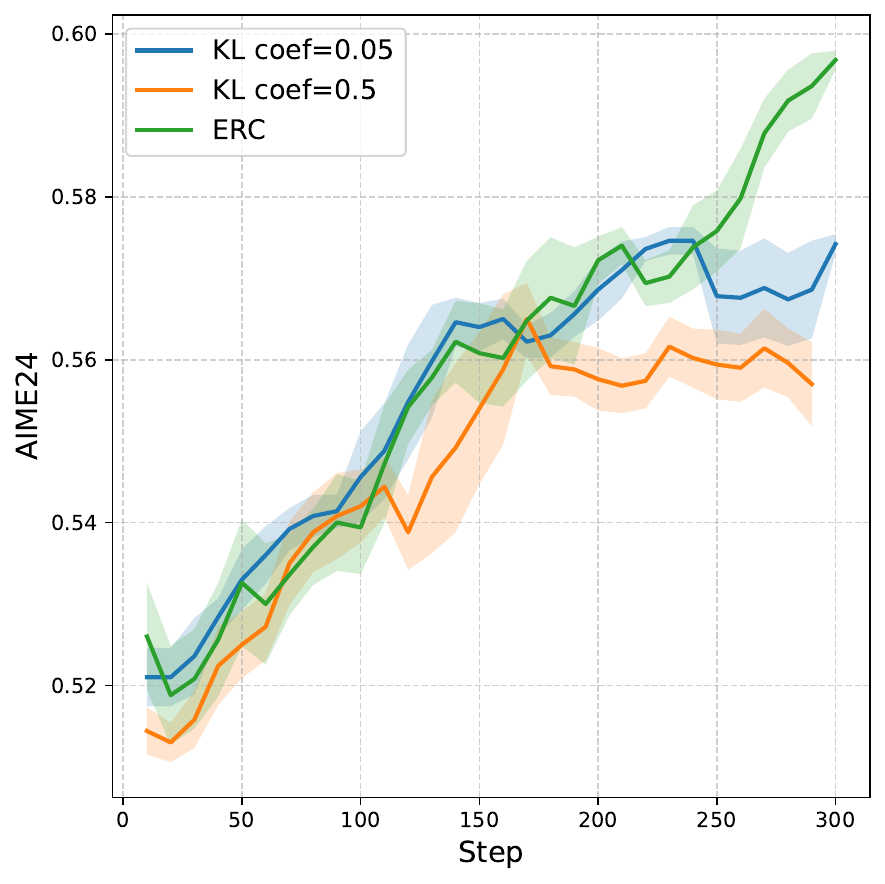}
        \caption{AIME24 Accuracy}
        \label{fig:kl_coef_aime24}
    \end{subfigure}
    \hfill
    \begin{subfigure}{0.48\linewidth}
        \centering
        \includegraphics[width=\linewidth]{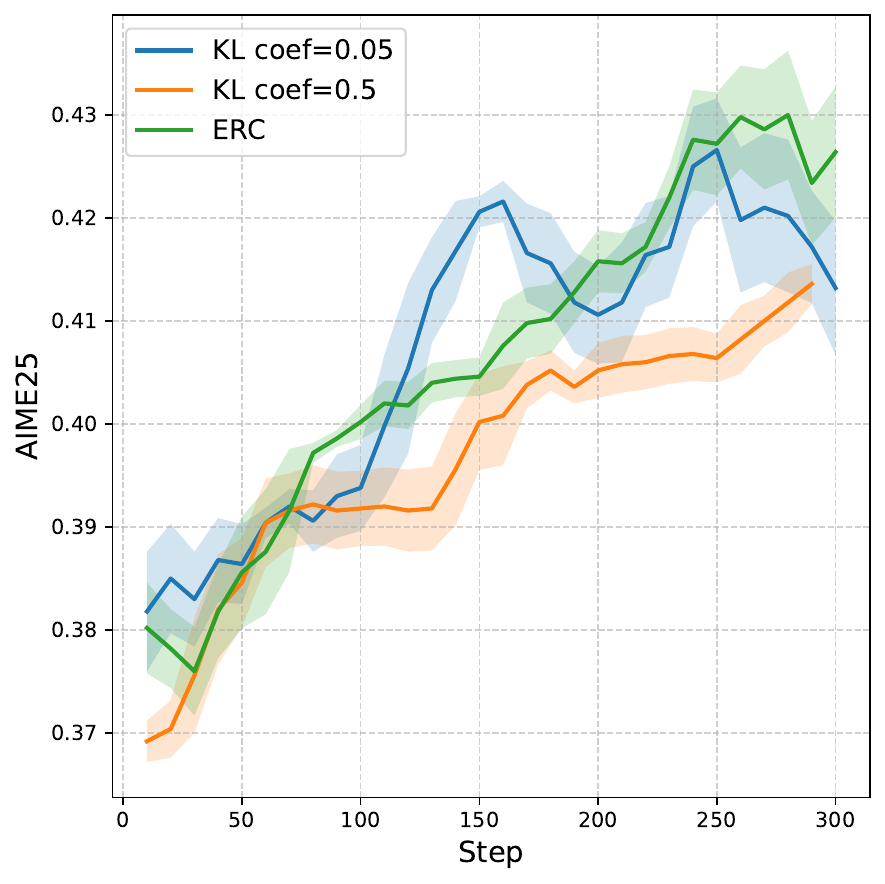}
        \caption{AIME25 Accuracy}
        \label{fig:kl_coef_aime25}
    \end{subfigure}
    \caption{Performance comparison of ERC and KL-regularized methods with varying coefficients. All methods are trained on the DS-R1-Distill-Qwen-7B model.}
\label{fig:kl_coef}
\end{figure}




To compare the performance of ERC with KL-regularization methods, we conducted evaluation on the AIME24 and AIME25 benchmarks. As shown in Figure \ref{fig:kl_coef}, ERC outperforms PPO-penalty (i.e., the KL-regularized approach) on both datasets.

Although both methods impose global constraints, their mechanisms differ fundamentally. KL divergence enforces a pointwise constraint, requiring the probability distributions of the old and new policies to remain close for every individual action. While this strict local regulation can stabilize training, it inevitably limits effective policy exploration, shrinking the update step sizes and making it harder for the model to escape local optima and reach higher-performing regions.

In contrast, ERC implements a distribution-level soft constraint. Rather than directly restricting the probability of each token, it monitors the evolution of the overall policy distribution via the entropy ratio. This mechanism selectively clips updates that significantly deviate from the trust region while preserving sufficient flexibility for exploration within reasonable bounds. Consequently, ERC encourages more efficient exploration while maintaining training stability, enabling the model to converge faster to superior performance.

\subsection{ERC vs. Entropy Regularization}



\begin{figure}[t]
    \centering
    \begin{subfigure}{0.48\linewidth}
        \centering
        \includegraphics[width=\linewidth]{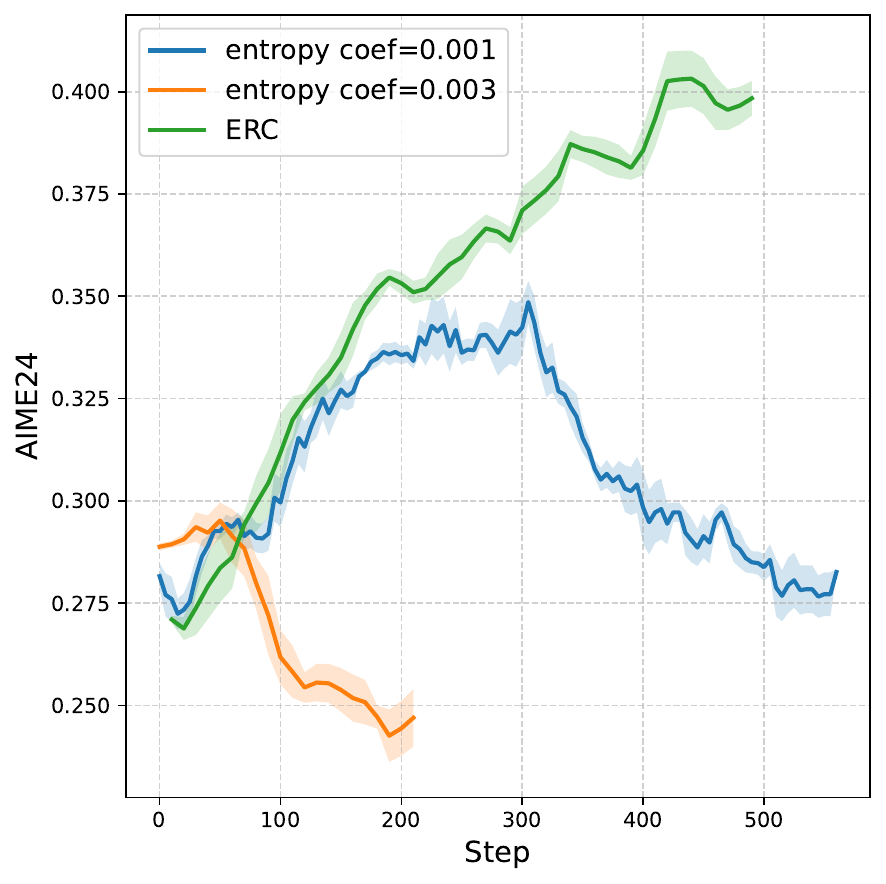}
        \caption{AIME24 Accuracy}
        \label{fig:entropy_coef_aime24}
    \end{subfigure}
    \hfill
    \begin{subfigure}{0.48\linewidth}
        \centering
        \includegraphics[width=\linewidth]{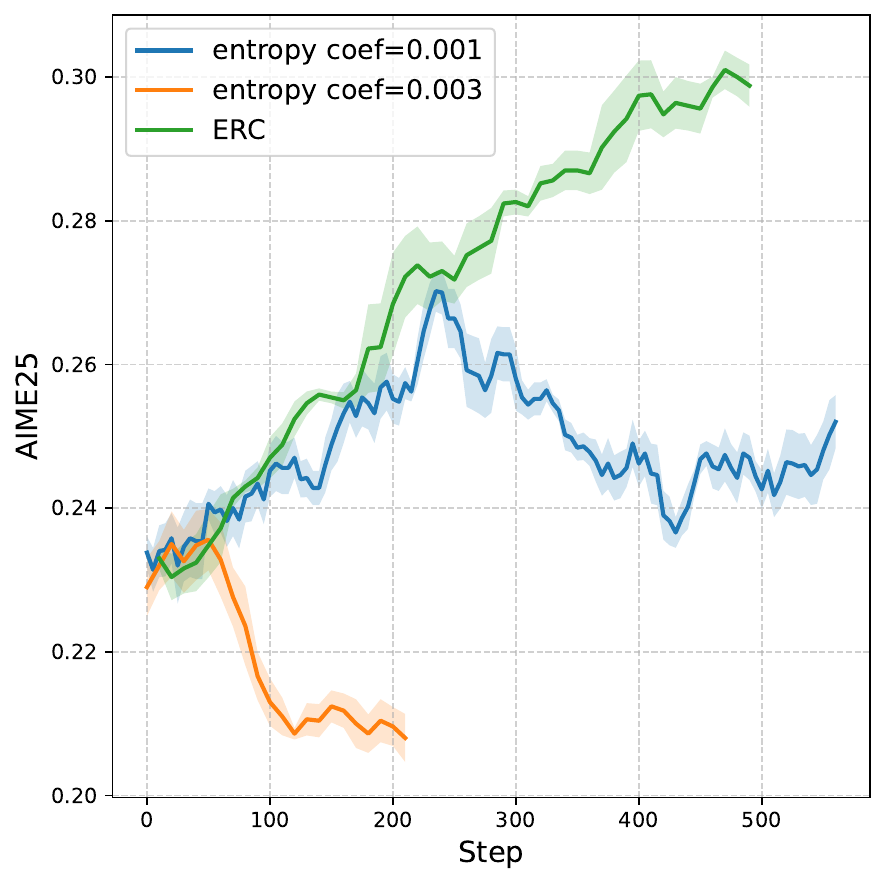}
        \caption{AIME25 Accuracy}
        \label{fig:entropy_coef_aime25}
    \end{subfigure}
    \caption{Performance comparison of ERC with entropy-regularized methods using different regularization coefficients. All methods are trained on the DS-R1-Distill-Qwen-1.5B model.}
\label{fig:entropy_coef}
\end{figure}

To compare the performance of erc with entropy regularization methods, we evaluated the method that directly incorporates entropy penalty during RL training on the aime24 and aime25 benchmarks.
As shown in Figure \ref{fig:entropy_coef}, ERC achieves significantly better performance. This advantage stems from a fundamental difference in how the two methods stabilize training through entropy: while entropy regularization can only mitigate unidirectional instability, ERC’s bidirectional clipping mechanism effectively addresses both directions of entropy fluctuations during policy evolution.

Specifically, entropy regularization adds an entropy term to the objective to encourage exploration and prevent premature entropy collapse. However, it provides limited control in the opposite scenario—entropy explosion—where the policy becomes excessively stochastic and exploration is no longer guided. As a result, the stability it ensures is inherently limited.

In contrast, ERC introduces entropy-ratio clipping with both lower and upper bounds. The lower bound prevents the policy from becoming overly conservative and collapsing into low-entropy regions, while the upper bound constrains overly aggressive updates that could lead to entropy explosion. This symmetric, bidirectional constraint ensures that the policy’s exploratory behavior evolves smoothly within a reasonable and controllable range, maintaining both stability and effective exploration.

\subsection{Comparison with Sequence-Level Clipping}

\begin{figure}[t]
    \centering
    \begin{subfigure}{0.48\linewidth}
        \centering
        \includegraphics[width=\linewidth]{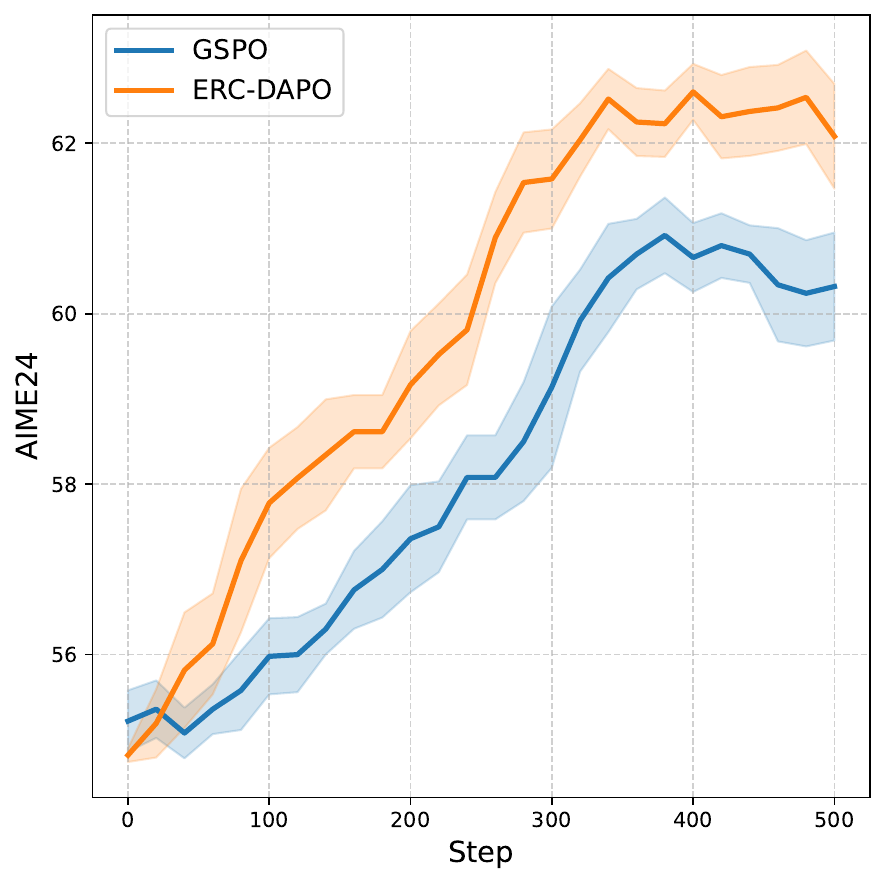}
        \caption{AIME24 Accuracy}
    \end{subfigure}
    \hfill
    \begin{subfigure}{0.48\linewidth}
        \centering
        \includegraphics[width=\linewidth]{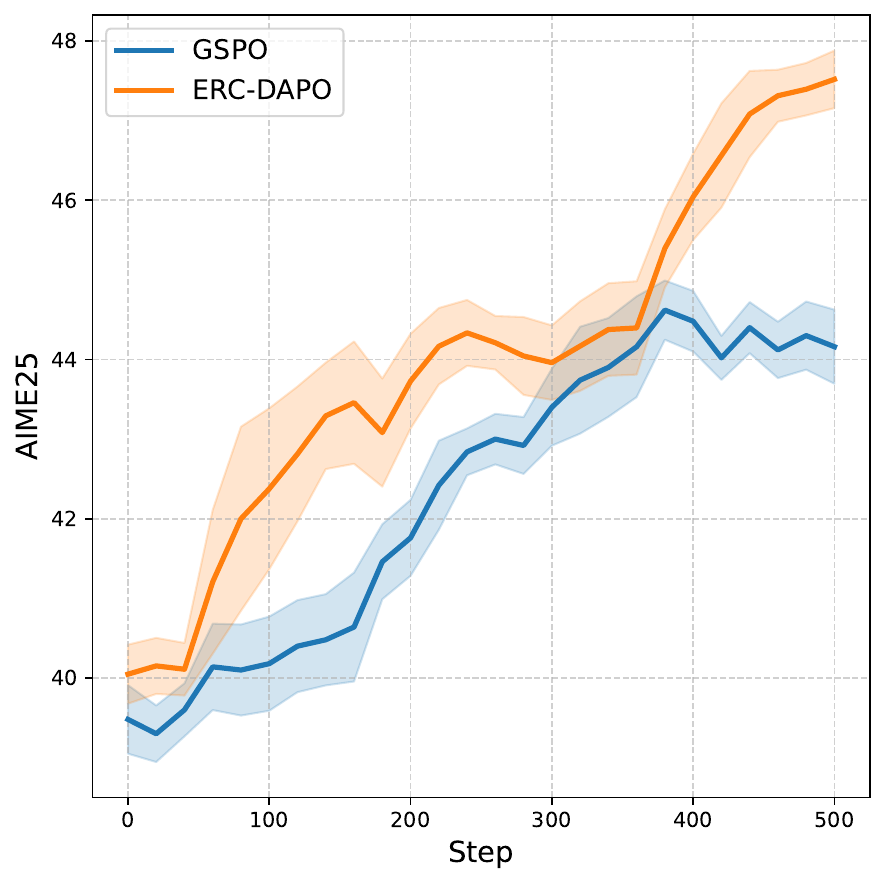}
        \caption{AIME25 Accuracy}
    \end{subfigure}
    \caption{Performance comparison of ERC with the sequence-level clipping method. All methods are trained on the DS-R1-Distill-Qwen-7B model.}
\label{fig:sequence_level}
\end{figure}

In this section, we compare ERC with a sequence-level clipping method~\citep{DBLP:journals/corr/abs-2507-18071}. Following the optimal configuration of GSPO~\citep{DBLP:journals/corr/abs-2507-18071}, we conducted experiments on DS-R1-Distill-Qwen-7B, where the average clipping ratio of tokens was approximately 15\%. As shown in Figure ~\ref{fig:sequence_level}, we present the metric trends for AIME24 and AIME25 during training. It can be observed that ERC-DAPO consistently demonstrates a clear advantage on both benchmarks. This indicates that the token-level clipping approach, which combines PPO-clip and ERC, still holds significant potential compared to sequence-level clipping. Additionally, it is worth noting that ERC and sequence-level clipping are orthogonal and can be used simultaneously.


\section{Conclusion}


Reinforcement learning for large language models has long suffered from training instability, primarily caused by trust-region deviation during optimization. Although PPO-clip mitigates part of this deviation, its fundamental limitation lies in only constraining the probability changes of sampled actions. Probability shifts among unsampled actions remain uncontrolled and can accumulate to cause significant trust-region drift. To address this issue, we propose using the entropy ratio between the new and old policies as a global measure of exploration change, and based on this, we design the ERC method. ERC imposes a bidirectional constraint on the global policy distribution, effectively alleviating trust-region deviation and stabilizing training. Experiments across multiple model scales demonstrate that ERC consistently outperforms baseline methods. Further empirical analysis shows that ERC not only suppresses trust-region drift and significantly enhances training stability, but also preserves the necessary exploratory behavior of the policy, ultimately improving final model performance.

\section*{Limitations}

Although the proposed ERC method demonstrates compelling results in mathematical reasoning tasks, its generalization to other domains, such as code generation or agent-based reinforcement learning, remains an open question due to computational constraints. We acknowledge that empirical validation across a broader range of domains would strengthen the claims regarding the method's universality. Therefore, extending ERC to these areas constitutes an important direction for our future work.

\bibliography{custom}

\clearpage

\end{document}